\documentclass[conference]{IEEEtran}
\IEEEoverridecommandlockouts

\usepackage{cite}
\usepackage{amsmath,amssymb,amsfonts}
\usepackage{algorithmic}
\usepackage{graphicx}
\graphicspath{{../}}
\usepackage{textcomp}
\usepackage{xcolor}
\usepackage{booktabs}
\usepackage{cuted}
\usepackage{caption}

\usepackage{pifont}
\newcommand{\method}{ART}

\definecolor{checkgreen}{HTML}{2E7D32}

\def\BibTeX{{\rm B\kern-.05em{\sc i\kern-.025em b}\kern-.08em
    T\kern-.1667em\lower.7ex\hbox{E}\kern-.125emX}}
\begin{document}

\title{Augmentation with Reinforcement Transformation for Low-textured Scientific Sketch Image Retrieval}

\author{
\IEEEauthorblockN{
Zhiqiang Wang\IEEEauthorrefmark{1},
Sheri L. Barton\IEEEauthorrefmark{2},
Jennifer H. Johnson\IEEEauthorrefmark{2},
Yi He\IEEEauthorrefmark{3}, and
Xingquan Zhu\IEEEauthorrefmark{1}
}
\IEEEauthorblockA{
\IEEEauthorrefmark{1}
Department of Electrical Engineering and Computer Science,
Florida Atlantic University, Boca Raton, USA\\
\IEEEauthorrefmark{2}
Mote Marine Laboratory, Sarasota, USA\\
\IEEEauthorrefmark{3}
Department of Data Science, William \& Mary,
Williamsburg, USA\\
\{zwang2022, xzhu3\}@fau.edu; \{sheri, jjohnson\}@mote.org; yihe@wm.edu
}
}

\maketitle

\begin{abstract}
Low-textured sketch images are widely used in scientific applications, including wildlife conservation, archaeological illustration, and forensic composites. Learning from and retrieving such images is challenging because they lack detailed surface textures and visual variations commonly exploited by modern deep learning models. This paper proposes Augmentation with Reinforcement Transformation (\method) to fine-tune pretrained models to learn effective representations for accurate sketch image retrieval. Our key observation is that sketches are particularly sensitive to image transformations, such as rotation, flipping, zooming, and stretching. While some transformations preserve their structural integrity, others may introduce distortions that hinder representation learning. Instead of randomly choosing transformations to construct fine-tuning views, \method\ formulates augmentation selection as a multi-armed bandit (MAB) problem, where transformations are treated as arms and their effectiveness is continuously evaluated through rewards. This enables \method\ to adaptively construct reward-guided view banks before self-supervised fine-tuning; the subsequent representation loss does not feed back to ART. On Florida manatee sketches provided by Mote Marine Laboratory, ART achieves 71.40\% average Top-1 accuracy for transformed-scar retrieval, compared with 68.83\% for random augmentation selection, and 55.84\% for location-aware sketch retrieval, compared with 18.18\% for frozen CLIP.
\end{abstract}

\begin{IEEEkeywords}
data augmentation, combinatorial bandits, low-textured images, sketch
retrieval, self-supervised representation learning
\end{IEEEkeywords}

\section{Introduction}
\label{sec:introduction}

Scientific databases often preserve observations in forms that differ sharply from the natural photographs used to pretrain modern vision models. Florida manatee photo-identification is one example. Accurately identifying individual manatees plays a vital role in conservation studies~\cite{wang2025manatee} by answering important questions such as whether this animal has been seen before, where it has been seen, and when it was seen.

\begin{figure}
    \centering
    \includegraphics[width=\linewidth]{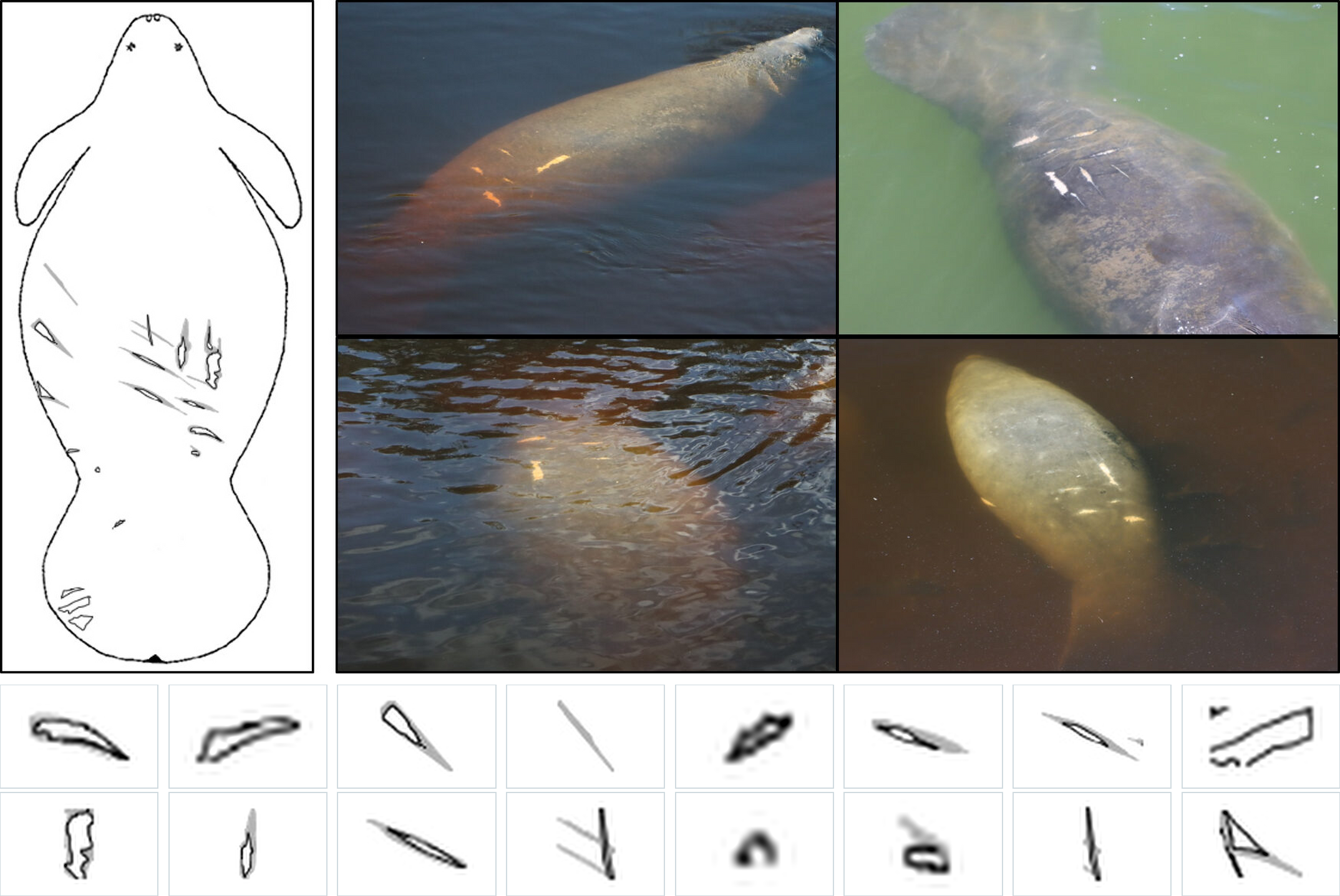}
    \caption {An example of using low-textured sketch image to help identify Florida manatee in conservation study. Top right shows four images of the same animal (DUSW11665) encountered/captured at different times/locations during Mote Marine Lab researchers' field study. Top left shows a hand drew sketch image with scars being marked to help identify the animal. Individual scars extracted from the sketch are shown at the bottom. }
    \label{fig:manatee_sketch_marker_examples}
\end{figure}

While individual manatee identification is important, using photos for identification is prohibitively difficult because such images are often taken in poor visibility, at arbitrary orientations, with distractions, and show only a small portion of the animals, as shown in Figure~\ref{fig:manatee_sketch_marker_examples}. As a result, for years, researchers have been using scars and mutilations to distinguish
animals and support longitudinal studies of survival, reproduction, and
population status \cite{beck1995automated}. 

In collaboration with the Mote Marine Laboratory, we have access to a real-world manatee conservation study database, where field observations are  transcribed as sparse line sketches
on a standardized body diagram. Identifying a newly observed animal, therefore
, requires searching a large historical sketch database for records with
corresponding scar evidence.

\begin{figure}
\centering
\includegraphics[width=\linewidth]{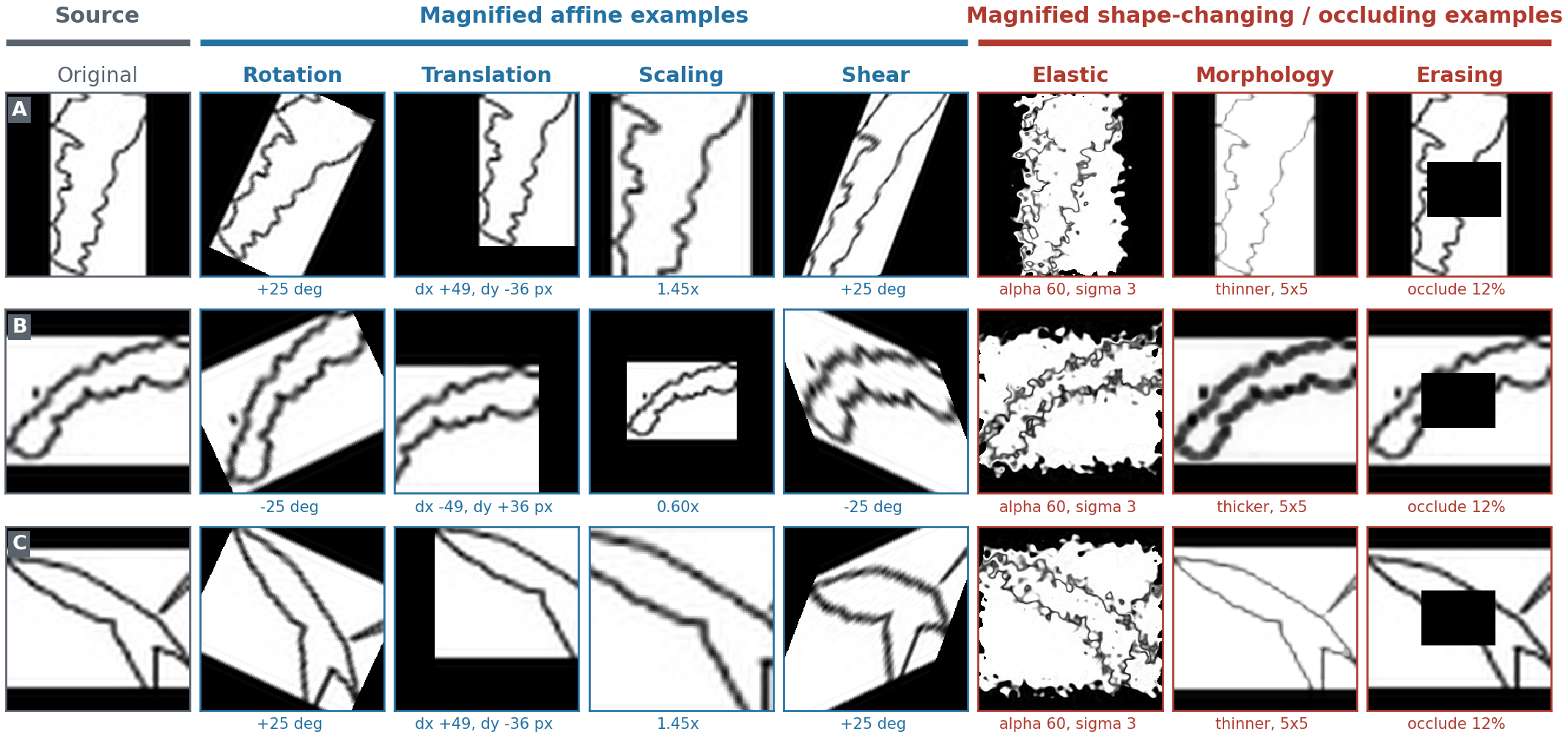}
\caption {Examples of manatee scars and results from seven different transformations, separated into two groups (affine \textit{vs.} shape transformations). Each row denotes one scar. From left to rights, first column denotes original scar, and remaining columns denote transformations. }
\label{fig:augmentation_examples}
\end{figure}

\begin{table*}
\caption{A summary and comparison of different representative augmentation designs, showing uniqueness of the proposed Augmentation with Reinforcement Transformation (ART), compared to other methods. \emph{Opt.} indicates explicit
optimization of exploitation, exploration, or transformation grouping. Grouping
means that transformation identities or combinations are optimized; it does not refer
to sample or class grouping. Complexity is qualitative because the methods use
different datasets and training budgets.}
\label{tab:augmentation_designs}
\centering
\small
\setlength{\tabcolsep}{2.5pt}
\newcommand{\arraystrech}{1.25}
\begin{tabular}{@{}p{3.8cm}p{4.45cm}cccp{3.6cm}p{2.5cm}@{}}
\toprule
Approach & Optimized unit & \shortstack{Opt.\\exploit.} &
\shortstack{Opt.\\explore} & \shortstack{Opt.\\grouping} & Feedback &
Complexity \\
\midrule
RandAugment \cite{cubuk2020randaugment}
  & Global count/magnitude & -- & -- & -- & Validation tuning & Low \\
AutoAugment \cite{cubuk2019autoaugment}
  & Ordered two-transformation subpolicies & \textcolor{green}{\checkmark} & \textcolor{green}{\checkmark} & \textcolor{green}{\checkmark}
  & Validation accuracy & Very high \\
SelfAug./TeachAug. \cite{reed2021selfaugment,suzuki2022teachaugment}
  & Policy under label-free or teacher evaluation & \textcolor{green}{\checkmark} & \textcolor{green}{\checkmark} & \textcolor{green}{\checkmark}
  & Proxy task/teacher signal & High \\
AdaAug/MADAug \cite{cheung2021adaaug,hou2023learn}
  & Class/instance/stage policy & \textcolor{green}{\checkmark} & -- & -- & Validation loss & High \\
BDA \cite{lu2023bda}
  & Transferable transformation-pair policies & \textcolor{green}{\checkmark} & \textcolor{green}{\checkmark} & \textcolor{green}{\checkmark}
  & Net validation prediction gain & Medium--high \\
EEA \cite{xiao2024data}
  & Magnitude at each augmentation depth & \textcolor{green}{\checkmark} & \textcolor{green}{\checkmark} & --
  & Correct-class probability & Medium \\
ADAMAB \cite{tang2026adaptive}
  & Class selected for synthesis & \textcolor{green}{\checkmark} & \textcolor{green}{\checkmark} & --
  & Gradient-shift acquisition & High \\
PDA/Data Reinf. \cite{yu2019pda,faghri2023reinforce}
  & Perturbation schedule/enriched examples & -- & -- & --
  & Adversarial/teacher signal & Medium--high \\
\textbf{ART (ours)}
  & \textbf{Two atomic image transformations} & \textcolor{green}{\checkmark} & \textcolor{green}{\checkmark} & \textcolor{green}{\checkmark}
  & Label-free sparse-structure reward & Low--medium \\
\bottomrule
\end{tabular}
\end{table*}

The discriminative evidence in these sketches is both local and extremely
sparse. An extracted scar has a median unpadded size of only
$16\times12$ pixels and contains almost no color or surface texture; its
identity is carried by a few strokes, junctions, and relative proportions. Figure \ref{fig:manatee_sketch_marker_examples} presents some samples.
Representation learning must consequently tolerate benign variation in drawing
and registration without becoming invariant to changes that alter the scar
itself. This makes data augmentation consequential rather than routine. Every
transformation asserts that the transformed view preserves the source identity,
yet elastic deformation, morphology, or erasing can move, thicken, or remove
precisely the stroke evidence needed for retrieval. Mild affine changes are
often safer, but even individually reasonable transformations can become destructive
when composed.

This asymmetry exposes a limitation of fixed augmentation policies. Uniformly
sampling a transformation library ignores whether earlier transformations preserved
the evidence of interest, while restricting the library by hand cannot adapt to
useful but less predictable combinations. The central question is therefore
not simply how much augmentation to apply, but which transformations should be
reinforced by observed outcomes and which should be suppressed.
Figure~\ref{fig:augmentation_examples} illustrates why adaptive selection is
preferable to classifying a transformation as universally safe or destructive.

We address this question with \emph{Augmentation with Reinforcement
Transformation} (ART), a feedback-guided augmentation framework for low-textured sketch images. A shared-credit combinatorial bandit controller
\cite{auer2002finite,chen2013combinatorial} uses a fixed, label-free structural
reward to reinforce transformations that produce useful, scar-preserving views
while continuing to explore uncertain choices. ART therefore requires neither
retrieval labels nor an end-to-end reinforcement-learning agent. We compare ART
with uniform sampling over the same transformation pairs, isolating adaptive
selection from the choice of augmentation library.

Across five locked folds and three optimization seeds, ART improves balanced
Top-1 on transformed-scar retrieval by 2.57 percentage points over the matched
uniform control, satisfying the prespecified strong-confirmation criterion.
This benchmark measures robustness to transformed instances, not generalization
to unseen animal identities or other low-texture domains. In a separate fixed
sketch-to-sketch experiment, location-aware ART scar matching achieves 55.84\%
Top-1, compared with 18.18\% for the strongest frozen whole-image encoder. 

The contributions are threefold:

\begin{itemize}
  \item We introduce label-free, feedback-guided augmentation for learning sparse
  scar representations, with shared credit across composed transformations.
  \item We isolate the benefit of adaptive selection through a matched uniform
  control and a multi-fold, multi-seed paired evaluation.
  \item We use the learned scar representations for location-aware sketch
  retrieval and evaluate retrieval accuracy and system scaling on the manatee
  database.
\end{itemize}



\section{Related Work}
\label{sec:related_work}

\subsection{Learned Data Augmentation}

Learned augmentation has progressively shifted the design question from
manually choosing transformations to defining what a policy should optimize.
AutoAugment searches ordered subpolicies with reinforcement learning, whereas
RandAugment compresses the search to a shared transformation count and magnitude
\cite{cubuk2019autoaugment,cubuk2020randaugment}. Subsequent work conditions
policies on classes, samples, models, or training stages. AdaAug and MADAug
learn adaptive policies, SelfAugment replaces label-based validation with
self-supervised evaluation, and TeachAugment uses a teacher to prevent
adversarial transformations from destroying recognizable content
\cite{cheung2021adaaug,reed2021selfaugment,suzuki2022teachaugment,hou2023learn}.
Progressive augmentation and dataset reinforcement instead regulate
perturbation difficulty or construct reusable enriched examples
\cite{yu2019pda,faghri2023reinforce}.

Bandit-based approaches are the closest methodological precedents for ART. BDA
uses validation gains to search transferable transformation-pair policies, while EEA
uses bandit feedback to adapt deformation magnitude but samples transformation
identity randomly \cite{lu2023bda,xiao2024data}. ADAMAB applies UCB to select
the class for which synthesized examples are acquired, rather than the
transformations applied to a source image \cite{tang2026adaptive}. These
methods establish feedback-guided augmentation and transformation composition as
prior ideas. ART addresses a narrower unresolved setting: selecting atomic
transformation combinations from label-free feedback designed to preserve the topology of
sparse scars. Table~\ref{tab:augmentation_designs} locates this distinction
in the optimized decision and feedback signal, rather than claiming novelty
from bandits alone.

\subsection{Sketch and Local-Evidence Retrieval}

Sketch-based image retrieval commonly learns a cross-modal embedding between a
freehand query sketch and natural photographs. Zero-shot systems emphasize
transfer to unseen semantic categories, while fine-grained systems retrieve a
particular photo instance and may reason over the order or completeness of user
strokes \cite{dey2019doodle,bhunia2020sketch,lin2023zero}. Large-scale image
retrieval also demonstrates the value of local rather than holistic descriptors
when discriminative evidence occupies only a small image region
\cite{noh2017large}.

Our setting differs in both modality and evidence organization. Queries and
database records are scientific sketches drawn on the same registered anatomical
template, not sketch--photo pairs. The shared body outline is visually dominant
but weakly discriminative, whereas identity evidence appears in a variable set
of small closed scars. We therefore retrieve through detected scars
and combine their learned appearance with registered-canvas location. The
reused detector and location-aware aggregation follow the prior manatee sketch
system \cite{wang2025manatee}; the present contribution concerns how reinforced
augmentation learns the scar representation used by that retrieval pipeline.
Our sketch-level experiment therefore compares this local-evidence design with
frozen holistic representations rather than with cross-modal sketch--photo
systems.

\begin{figure*}[t]
\centering
\includegraphics[width=1\linewidth]{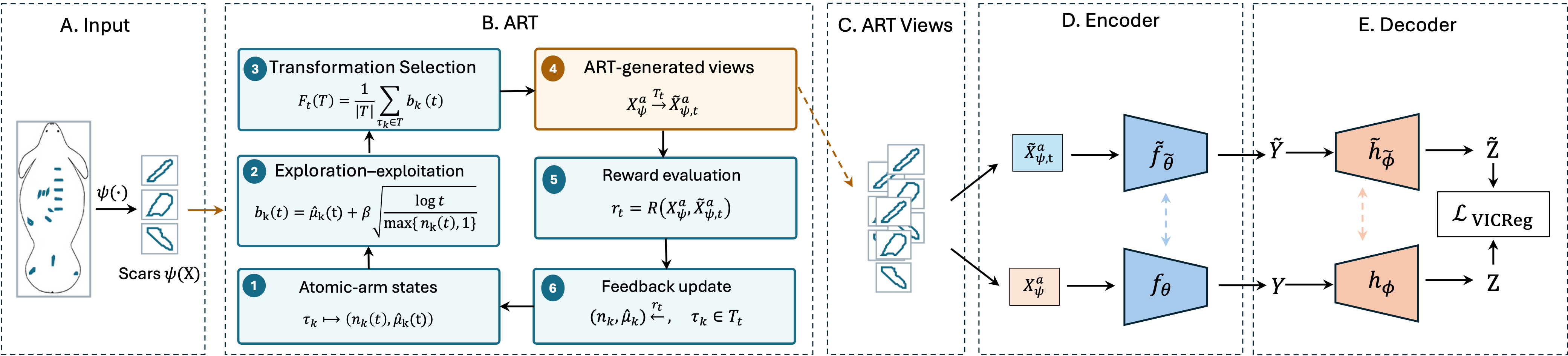}
\caption{ART--VICReg fine-tuning workflow. (A--C) Detected scars enter the
six-step ART loop, which selects transformation combinations, evaluates the
resulting views, updates shared atomic-arm states, and caches ART-generated
views. (D--E) Two views of the same scar (a source--transformed pair is shown)
pass through parameter-sharing encoder and training-only expander branches
under the VICReg objective. VICReg updates $f_\theta$ and $h_\phi$ but does not
feed back to ART; $h_\phi$ is discarded after training and
$f_{\theta^\star}$ is retained for retrieval.}
\label{fig:art_reinforcement}
\end{figure*}

\section{Problem Definition}
\label{sec:problem_definition}
This section first summarizes key symbols and notations, then defines two research problems we intend to address. 



\subsection{Key Symbols and Notations}

\begin{center}
\begin{minipage}{\columnwidth}
\centering
\footnotesize
\setlength{\tabcolsep}{3pt}
\renewcommand{\arraystretch}{1.2}
\begin{tabular}{@{}p{0.31\columnwidth}p{0.63\columnwidth}@{}}
\toprule
Symbol & Meaning \\
\midrule
\multicolumn{2}{@{}l}{\emph{Sketches and scars}} \\
$X,Q,\mathcal D=\{X_j\}_{j=1}^{n_{\mathcal D}}$ & A sketch image $X$, a
query sketch $Q$, and a sketch database $\mathcal D$ containing $n_{\mathcal D}$
sketches. \\
$\psi(\cdot),\psi(X),n_{\psi(X)}$ & The scar detector $\psi(\cdot)$, the set
of cropped scars $\psi(X)$ extracted from $X$, and the scar count
$n_{\psi(X)}=|\psi(X)|$. \\
$X_{\psi}^{a}$ & The $a$th cropped scar image of sketch $X$, so that
$\psi(X)=\{X_{\psi}^{a}\}_{a=1}^{n_{\psi(X)}}$. \\
\addlinespace[2pt]
\multicolumn{2}{@{}l}{\emph{Reinforcement-guided transformation selection}} \\
$\mathcal T=\{\tau_k\}_{k=1}^{K},\mathcal A$ & The available augmentation
transformations and the permitted combinations of those transformations. \\
$T_t\in\mathcal A$ & The transformation combination selected by ART at round
$t$. In the overall method flow, $T_t$ is treated as one transformation. \\
$\widetilde X_{\psi,t}^{a}$ & The transformed scar image produced at round
$t$: $\widetilde X_{\psi,t}^{a}=T_t(X_{\psi}^{a})$. \\
$r_t$ & The reward evaluating whether $\widetilde X_{\psi,t}^{a}$ is a useful
and scar-preserving augmentation. Rewards guide later transformation
selections. \\
\addlinespace[2pt]
\multicolumn{2}{@{}l}{\emph{Scar representation and sketch retrieval}} \\
$f_{\theta},f_{\theta^\star}$ & The retained scar-representation path before
and after VICReg fine-tuning; its pretrained visual backbone remains fixed. \\
$h_\phi$ & The training-only VICReg expander, optimized jointly with the
trainable part of $f_\theta$ and discarded after fine-tuning. \\
$\mathbf z_X^a$ & The embedding of scar $X_{\psi}^{a}$ produced by the
fine-tuned model: $\mathbf z_X^a=f_{\theta^\star}(X_{\psi}^{a})$. \\
$S(Q,X_j)$ & The selected overall similarity between query sketch $Q$ and
database sketch $X_j$. \\
\\
\bottomrule
\end{tabular}
\end{minipage}
\end{center}

\subsection{Problem Formulation}
Given a query sketch $Q$, the overall goal of our research is to retrieve from a low-textured scientific sketch image database $\mathcal{D}$ with the best retrieval accuracy. We address this task through two separate objectives: (1) learn good embedding features to represent each scar; and (2) use the learned scar embedding features to help retrieve the best matching sketch images. In the following, we formulate two research problems to achieve these two objectives

\subsubsection{Augmentation with Reinforcement Transformation Objective}

Given a low-textured scientific sketch image database $\mathcal{D}$, the ART learning \textbf{aims} to apply reinforcement transformations to augment scars to effectively train a scar encoder $f_{\theta^\star}$ that maps
each detected scar $X_{\psi}^{a}$ to an embedding feature.

\subsubsection{Scar-based Sketch Retrieval Objective}
By using learned scar embedding features $X_{\psi}^{a}$, the scar-based sketch retrieval \textbf{aims} to search for the best matching sketch image to the query sketch $Q$.


During retrieval, scar embeddings from a query sketch are compared with those
from the database sketches, and the resulting evidence is aggregated to rank
the database records. For a newly recorded sketch, this ranking helps users
determine whether the observed manatee may correspond to an existing
photo-identification record.

\section{Augmentation with Reinforcement Transformation}
\label{sec:art_method}

Augmentation with Reinforcement Transformation (ART) formulates data
augmentation as a sequential decision problem. Figure~\ref{fig:art_reinforcement}
summarizes the complete training workflow. After scar extraction (A), ART uses
its six-step selection-and-feedback loop to generate and cache transformed
views (B--C). VICReg then processes two same-scar views through shared encoder
and training-only expander branches (D--E). The workflow is sequential and
one-way: VICReg gradients and retrieval outcomes are not returned to the ART
controller.

Given a transformation library
$\mathcal T=\{\tau_k\}_{k=1}^{K}$ and a set $\mathcal A$ of permitted
transformation combinations, ART selects one combination $T_t\in\mathcal A$ at
round $t$. Unlike a fixed or randomly sampled augmentation policy, this choice
is guided by rewards observed in previous rounds. The reinforcement mechanism
therefore balances two objectives: exploiting transformations that have
produced useful training examples and exploring transformations whose effects
remain uncertain.

For each transformation $\tau_k$, let $n_k(t)$ be its number of previous
selections and let $\widehat\mu_k(t)$ be its mean observed reward. ART assigns
the transformation the utility
\begin{equation}
b_k(t)=\widehat\mu_k(t)+\beta
\sqrt{\frac{\log t}{\max\{n_k(t),1\}}},
\label{eq:simplified_art_utility}
\end{equation}
where the first term represents past effectiveness and the second encourages
exploration. The parameter $\beta$ controls their balance.

For a candidate $T\in\mathcal A$, the utilities of its constituent
transformations are aggregated as
\begin{equation}
F_t(T)=\frac{1}{|T|}\sum_{\tau_k\in T}b_k(t),
\label{eq:simplified_art_score}
\end{equation}
where $T$ is a set of elementary transformations and $|T|$ is its cardinality,
that is, the number of transformations it contains. ART obtains the next
transformation by
\begin{equation}
T_t=\underset{T\in\mathcal A}{\arg\max}\;F_t(T).
\label{eq:simplified_art_selection}
\end{equation}

The selected $T_t$ is applied to a training scar to obtain
$\widetilde X_{\psi,t}^{a}=T_t(X_{\psi}^{a})$. ART compares this transformed
scar with its source $X_{\psi}^{a}$ through four components: structural
preservation $P_t$, frozen-feature consistency $C_t$, useful variation $V_t$,
and transformation harm $H_t$. All four quantities lie in $[0,1]$, and
$[u]_0^1=\min\{1,\max\{0,u\}\}$ denotes clipping to this interval.

Let $M$ and $\widetilde M_t$ be the source and transformed foreground masks,
$I_t$ their mask intersection over union, $g(\cdot)$ the fixed,
$\ell_2$-normalized CLIP encoder, and $\boldsymbol{\eta}_t$ the parameters of
$T_t$. The four reward components are written compactly as
\begin{equation}
\begin{aligned}
P_t &= \mathcal P(M,\widetilde M_t),\\
C_t &= [g(X_{\psi}^{a})^{\mathsf T}
g(\widetilde X_{\psi,t}^{a})]_0^1,\\
V_t &= \mathcal V(1-C_t,1-I_t),\\
H_t &= \mathcal H(\boldsymbol{\eta}_t).
\end{aligned}
\label{eq:art_reward_components}
\end{equation}
Here, $\mathcal P$ summarizes mask overlap, stroke proximity, and area
stability; $\mathcal V$ favors moderate feature and mask changes; and
$\mathcal H$ maps the sampled transformation type and magnitude to deformation
risk. These deterministic component functions and their internal settings are
fixed before evaluation and are not optimized during training. If a reliable
foreground mask cannot be extracted, $P_t$ is estimated from transformation
harm and $V_t$ from feature change only.

The four components are combined through nonnegative reward weights where $\sum_{i=1}^{4}w_i=1$ and $w_i\geq0$:
\begin{equation}
\begin{aligned}
r_t&=R(X_{\psi}^{a},\widetilde X_{\psi,t}^{a})\\
&=w_1P_t+w_2C_t+w_3V_t+w_4(1-H_t),
\end{aligned}
\label{eq:simplified_art_feedback}
\end{equation}

\begin{figure*}[t]
\centering
\includegraphics[width=\linewidth]{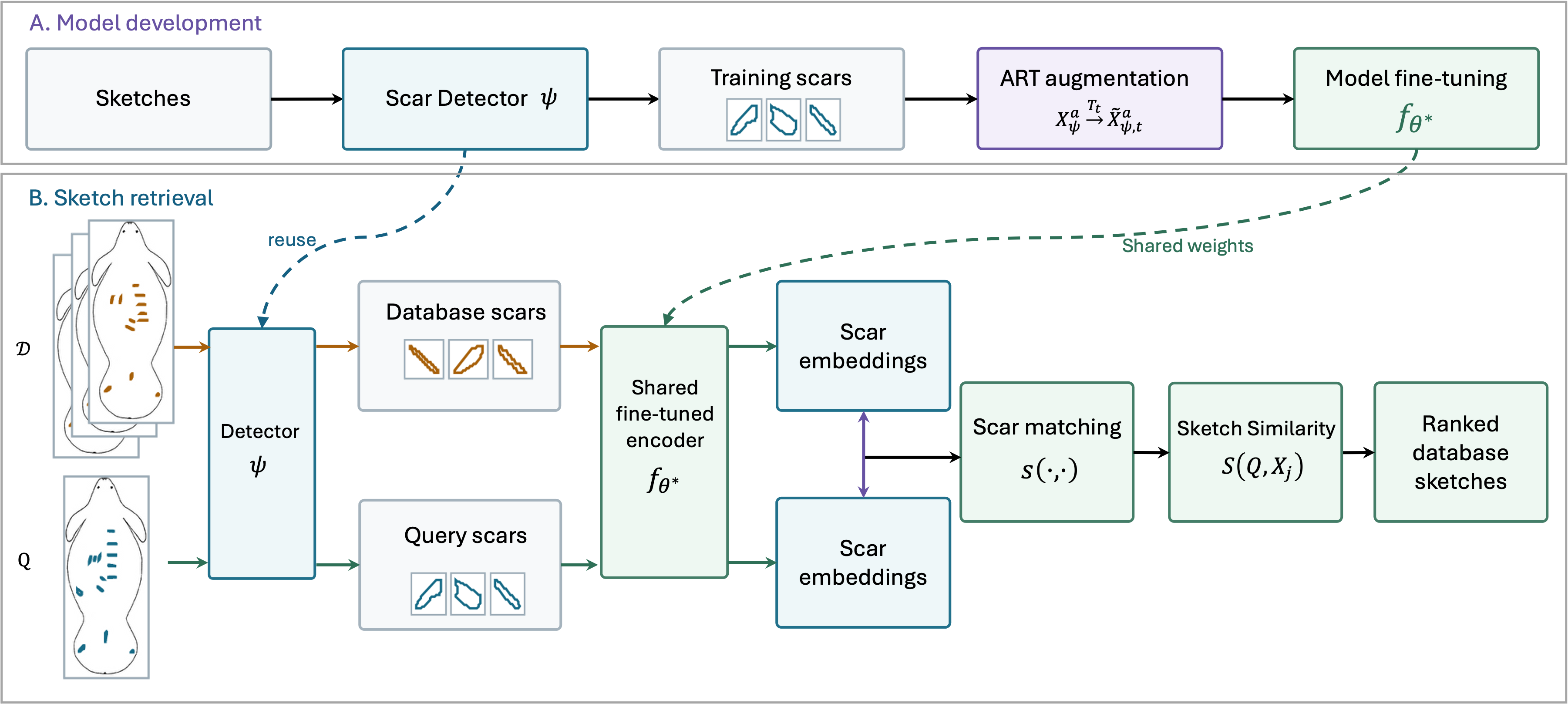}
\caption{Overview of the proposed framework. During model development, ART
constructs reward-guided scar-view banks that are subsequently used for VICReg
fine-tuning of $f_{\theta^\star}$. During retrieval, the shared retained encoder
produces query and database scar embeddings, which are matched and aggregated
to rank the database sketches.}
\label{fig:art_retrieval_system}
\end{figure*}

These are reward components, not losses optimized by backpropagation. The observed $r_t$
is credited to every transformation in $T_t$,
updating their estimated effectiveness before the next ART round. Repeating
this loop produces a bank containing the source scar and its ART-generated
views. The fixed visual backbone features for this bank are cached before
representation training begins.

For each VICReg training sample, two distinct views of the same scar are drawn
from the cached bank without replacement. Consequently, a pair may contain the
source and one transformed view, as illustrated in
Figure~\ref{fig:art_reinforcement}, or two different transformed views. The two
branches compute
$Y=f_\theta(V_i)$, $\widetilde Y=f_\theta(V_j)$,
$Z=h_\phi(Y)$, and $\widetilde Z=h_\phi(\widetilde Y)$ for sampled views
$V_i$ and $V_j$. Although the figure follows the original VICReg visual
convention by using tilded symbols in the upper branch, the branches are
Siamese and share parameters ($\widetilde\theta=\theta$ and
$\widetilde\phi=\phi$). Minimizing $\mathcal L_{\mathrm{VICReg}}$ updates
the trainable parameters of both the retained representation path $f_\theta$
and the training-only expander $h_\phi$; the pretrained visual backbone remains
fixed. After fine-tuning, $h_\phi$ is discarded and the resulting
$f_{\theta^\star}$ is passed to the retrieval stage.

\section{Scar-Based Sketch Retrieval}
\label{sec:simplified_scar_retrieval}

Figure~\ref{fig:art_retrieval_system} presents the framework as two connected
stages. The upper stage learns an encoder for individual scars. The lower stage
uses the resulting scar embeddings as the evidence for sketch retrieval.

\subsection{Scar Embedding Learning}

As described in Section~\ref{sec:art_method}, ART-guided VICReg fine-tuning
yields the retained scar encoder $f_{\theta^\star}$. For any detected scar
$X_{\psi}^{a}$, this encoder produces
\begin{equation}
\mathbf z_X^a=f_{\theta^\star}(X_{\psi}^{a}),
\label{eq:scar_embedding}
\end{equation}
where $\mathbf z_X^a$ is the scar representation used for retrieval.

\subsection{Sketch Retrieval from Scar Embeddings}

The retrieval stage consumes the scar embeddings defined in
Eq.~\eqref{eq:scar_embedding}. The same detector $\psi$ extracts scars from the
query sketch $Q$ and each database sketch $X_j$, and the shared
$f_{\theta^\star}$ maps them into the same representation space. This produces
query embeddings $\{\mathbf z_Q^a\}$ and database embeddings
$\{\mathbf z_{X_j}^b\}$ for direct comparison. ART is used only during model
development and is not applied to retrieval inputs.

Scar matching proceeds from local correspondences to a sketch-level score. We
evaluate two aggregations. The best-pair score retains the single strongest
correspondence between the two sketches:
\begin{equation}
S_{\max}(Q,X_j)=
\max_{\substack{1\leq a\leq n_{\psi(Q)}\\
1\leq b\leq n_{\psi(X_j)}}}s_{ab}(Q,X_j).
\label{eq:best_pair_similarity}
\end{equation}
The query-directed-mean score instead finds the best database scar for every
query scar and then averages those local matches:
\begin{equation}
\bar{S}_{\rightarrow}(Q,X_j)=\frac{1}{n_{\psi(Q)}}
\sum_{a=1}^{n_{\psi(Q)}}
\max_{1\leq b\leq n_{\psi(X_j)}}
s_{ab}(Q,X_j).
\label{eq:simplified_sketch_similarity}
\end{equation}
Here, $s_{ab}(Q,X_j)$ is either the appearance score alone or its fusion with
relative scar position on the standardized sketch template. For candidate sketch \(X_j\), \(S_{\max}\), $S_{max}$ uses the single highest-scoring pair among all query–candidate scar pairs. In contrast, \(\bar{S}_{\rightarrow}\) finds the best-matching scar in \(X_j\) for each query scar and averages these query-specific scores.

\section{Experimental Setup}
\label{sec:experimental_setup}

\begin{table}[h]
\caption{Sketch and scar level retrieval dataset}
\label{tab:dataset_summary}
\centering
\small
\setlength{\tabcolsep}{20pt}
\renewcommand{\arraystretch}{1.15}
\begin{tabular}{@{}lrrr@{}}
\toprule
Collection & Sketches & Identities & Scars \\
\midrule
Queries & 77 & 52 & 663 \\
Database & 1,645 & 1,635 & 12,221 \\
\bottomrule
\end{tabular}
\end{table}

\subsection{Dataset and Preprocessing}
\label{sec:dataset_construction}

Mote Marine Laboratory supplied hand-recorded Florida manatee scar patterns on
a standardized body diagram. We reuse the Mask R-CNN scar detector from prior
work~\cite{wang2025manatee}; detector training is not part of this study. The
same detector and preprocessing are used for model development, database
construction, and query processing.

Table~\ref{tab:dataset_summary} summarizes the query collection and the full
database. The fixed sketch-level experiment uses all 77 query sketches from 52
identities and a fixed gallery of 979 training sketches. Under the
identity-prefix rule, each query has exactly one matching gallery sketch. The
full database contains 1,645 sketches from 1,635 identities and supplies the
12,221 detected scars and the subsets used for system evaluation.

\subsection{Evaluation Protocols}
\label{sec:evaluation_protocols}

\paragraph{Transformed-scar benchmark}
The 12,221 source scars are partitioned into five fixed folds. In each fold,
1,000 training scars are used for self-supervised representation learning, and
held-out scars are used to construct the retrieval evaluation. Each query has
ten corresponding variants in each of three transformed test sets:
elastic-only, safe-affine, and mixed transformations. Top-1 retrieval is
correct when the first result is a transformed variant of the same source
scar. This benchmark evaluates robustness to transformed scar instances.

\paragraph{Fixed sketch-level identity retrieval}
Each query is ranked against the 979-sketch gallery. A retrieval is correct at
Top-$k$ when the unique identity-matched gallery sketch has rank at most $k$.
We report Top-1, Top-5, Top-10, Top-20, and mean reciprocal rank (MRR), retaining
all queries in the fixed evaluation manifest.

\paragraph{System evaluation}
The 77 searchable query sketches are also evaluated against databases of 100,
250, 500, 1,000, and all 1,635 searchable identities. We report database
construction time, retrieval latency, and memory use. This evaluation measures
computational scalability rather than retrieval accuracy or generalization to
unseen identities.

\subsection{Model and Training Configuration}
\label{sec:model_training}

All learned scar encoders use the ART--VICReg architecture described in
Section~\ref{sec:art_method}: a fixed pretrained CLIP ViT-B/32 visual backbone
\cite{radford2021learning}, a trainable three-layer representation head
$f_\theta$, and the training-only expander $h_\phi$. For each fold, VICReg
\cite{bardes2021vicreg} is trained for 50 epochs on 1,000 scars with batch size
16 and learning rate $10^{-4}$. Each scar has a 10-view cache containing the
source view and nine transformed views.

ART uses eight atomic transformations: small rotation, translation, scaling,
and shear, together with mild stroke noise, elastic deformation, morphology,
and erasing. All 28 unordered transformation pairs are eligible. We set the
exploration coefficient to $\beta=0.7$, the safety penalty to $0.05$, and the
risk-rejection threshold to $0.85$. The structural-reward weights are
$(w_1,w_2,w_3,w_4)=(0.35,0.25,0.20,0.20)$. These settings are fixed before
evaluation and do not use identity labels or retrieval outcomes.

\subsection{Sketch-Level Scoring and Whole-Image Baselines}
\label{sec:sketch_level_scoring}

The ART sketch-level scores reuse the five locked seed-42 fold encoders and
their precomputed detector crops. For each query scar and encoder,
we compute cosine similarity to every one of the eligible gallery scars
and standardize that complete score row. Averaging the five standardized rows
gives the appearance score $A_{ab}(Q,X_j)$. We similarly standardize the
negative normalized-centroid distance from a query scar to every eligible
gallery scar to obtain the location score $L_{ab}(Q,X_j)$. The location-aware
pair score follows the existing manatee retrieval rule:
\begin{equation}
F_{ab}(Q,X_j)=0.4A_{ab}(Q,X_j)+0.6L_{ab}(Q,X_j).
\label{eq:location_fusion}
\end{equation}
We use either $s_{ab}=A_{ab}$ or $s_{ab}=F_{ab}$ in the best-pair and
query-directed-mean aggregations of Eqs.~\eqref{eq:best_pair_similarity}
and~\eqref{eq:simplified_sketch_similarity}, yielding four ART variants.

Three frozen encoders provide holistic controls: ResNet50 with ImageNet-1K V2
weights and its 2,048-dimensional pooled representation
\cite{he2016deep}, OpenAI CLIP ViT-B/32 with its 512-dimensional image
representation \cite{radford2021learning}, and \texttt{facebook/dinov2-base}
with its 768-dimensional CLS-token representation \cite{oquab2023dinov2}. Embeddings are
$\ell_2$-normalized, and the gallery is ranked by cosine similarity.

\subsection{Comparisons and Statistical Reporting}
\label{sec:baselines_statistics}

We compare ART with uniform selection over the same 28 transformation
combinations, a fixed conservative augmentation strategy, RandAugment
\cite{cubuk2020randaugment}, and a conventional augmentation strategy. All
methods use the same fixed folds, training scars, embedding architecture,
10-view cache, and optimization budget. Uniform selection is the primary
control because it uses the same transformation space as ART without
reward-guided selection or the fixed safety adjustment.

The transformed-scar experiments use five folds and three optimization seeds
(42, 142, and 242), giving 15 matched fold--seed runs per method. Balanced
Top-1 is the unweighted mean across the three transformed test sets. The paired
ART-minus-uniform contrast uses a 10,000-replicate hierarchical bootstrap that
resamples folds and then seeds within folds. Reward-component removals use seed
42 over the same five folds and are treated as diagnostic ablations. The
sketch-level experiment is a deterministic, single-split comparison.

\section{Results}
\label{sec:results}

\begin{table*}[t]
\caption{Confirmatory transformed-scar Top-1 (\%). Each cell reports the mean on the first line and its hierarchical-bootstrap 95\% CI in parentheses on the second line, using 15 matched fold--seed cells. Balanced is the within-cell unweighted mean over the three transformed test sets. Bold denotes the largest mean in a column.}
\label{tab:confirmatory_core}
\centering
\small
\setlength{\tabcolsep}{25pt}
\renewcommand{\arraystretch}{2.0}

\begin{tabular}{@{}lrrrr@{}}
\toprule
Method & Elastic 10$\times$ & Safe 10$\times$ & Full 10$\times$ & Balanced \\
\midrule

ART shared-credit UCB &
\shortstack{$\mathbf{66.64}$\\{\scriptsize $(63.91\text{--}69.07)$}} &
\shortstack{$83.80$\\{\scriptsize $(82.40\text{--}85.42)$}} &
\shortstack{$63.76$\\{\scriptsize $(62.71\text{--}64.84)$}} &
\shortstack{$\mathbf{71.40}$\\{\scriptsize $(70.00\text{--}72.81)$}} \\

Uniform over identical 28 pairs &
\shortstack{$65.18$\\{\scriptsize $(62.53\text{--}67.31)$}} &
\shortstack{$81.09$\\{\scriptsize $(79.44\text{--}82.67)$}} &
\shortstack{$60.22$\\{\scriptsize $(58.62\text{--}62.00)$}} &
\shortstack{$68.83$\\{\scriptsize $(67.50\text{--}70.36)$}} \\

Fixed conservative augmentation &
\shortstack{$54.91$\\{\scriptsize $(52.18\text{--}57.69)$}} &
\shortstack{$\mathbf{88.13}$\\{\scriptsize $(86.18\text{--}90.07)$}} &
\shortstack{$\mathbf{68.11}$\\{\scriptsize $(66.00\text{--}70.42)$}} &
\shortstack{$70.39$\\{\scriptsize $(68.51\text{--}72.44)$}} \\

RandAugment ($N=2$, $M=9$) &
\shortstack{$44.71$\\{\scriptsize $(42.22\text{--}47.11)$}} &
\shortstack{$74.40$\\{\scriptsize $(72.09\text{--}76.62)$}} &
\shortstack{$55.29$\\{\scriptsize $(53.13\text{--}57.47)$}} &
\shortstack{$58.13$\\{\scriptsize $(56.01\text{--}60.21)$}} \\

Conventional full augmentation &
\shortstack{$43.11$\\{\scriptsize $(39.67\text{--}46.80)$}} &
\shortstack{$75.93$\\{\scriptsize $(73.98\text{--}78.51)$}} &
\shortstack{$61.89$\\{\scriptsize $(59.09\text{--}64.53)$}} &
\shortstack{$60.31$\\{\scriptsize $(58.06\text{--}62.99)$}} \\

\bottomrule
\end{tabular}
\end{table*}

\begin{table}[t]
\caption{Primary comparison on the identical 28-combination transformation space. Estimates and hierarchical-bootstrap 95\% confidence intervals (CIs) are in percentage points.}
\label{tab:art_uniform_primary}
\centering
\small
\setlength{\tabcolsep}{11pt}
\renewcommand{\arraystretch}{1.15}
\begin{tabular}{@{}lrr@{}}
\toprule
Quantity & Estimate & 95\% CI \\
\midrule
ART shared-credit UCB & $71.40$ & $70.00\text{ to }72.81$ \\
Uniform over identical 28 pairs & $68.83$ & $67.50\text{ to }70.36$ \\
Paired ART $-$ uniform & $2.57$ & $1.76\text{ to }3.32$ \\
\bottomrule
\end{tabular}
\end{table}

\begin{table*}[t]
\caption{Fixed sketch-level retrieval.
Top-$k$ entries are accuracy (\%) over the 77 scar-eligible queries; MRR is
reported on $[0,1]$. Bold denotes the largest value in each column.}
\label{tab:sketch_level_retrieval_77}
\centering
\small
\setlength{\tabcolsep}{20pt}
\renewcommand{\arraystretch}{1.35}
\begin{tabular}{@{}lrrrrr@{}}
\toprule
Method & Top-1 & Top-5 & Top-10 & Top-20 & MRR \\
\midrule
ResNet50
  & 14.29 & 22.08 & 25.97 & 35.06 & 0.1910 \\
CLIP ViT-B/32
  & 18.18 & 29.87 & 35.06 & 42.86 & 0.2400 \\
DINOv2-base
  & 16.88 & 32.47 & 37.66 & 41.56 & 0.2481 \\
\midrule
ART best pair
  & 14.29 & 25.97 & 32.47 & 38.96 & 0.2006 \\
ART mean pairs
  & 24.68 & 37.66 & 45.45 & 51.95 & 0.3172 \\
ART best pair + location
  & 35.06 & 46.75 & 54.55 & 70.13 & 0.4222 \\
ART mean pairs + location
  & \textbf{55.84}
  & \textbf{68.83}
  & \textbf{72.73}
  & \textbf{75.32}
  & \textbf{0.6193} \\
\bottomrule
\end{tabular}
\label{tab:sketch_level_retrieval}
\end{table*}

\begin{table*}[t]
\caption{Reward-component ablations at optimization seed 42 over five frozen folds. Balanced Top-1 is mean $\pm$ standard error; the last columns give the paired ablation-minus-full difference and fold-bootstrap 95\% CI.}
\label{tab:reward_ablation}
\centering
\small
\setlength{\tabcolsep}{14pt}
\renewcommand{\arraystretch}{1.3}
\begin{tabular}{@{}lrrr@{}}
\toprule
Reward & Balanced Top-1 & $\Delta$ & 95\% CI \\
\midrule
Full ART reward & $72.36\pm0.84$ & --- & --- \\
Without $P_t$ & $71.11\pm1.08$ & $-1.24$ & $-2.00\text{ to }-0.42$ \\
Without $C_t$ & $70.24\pm1.08$ & $-2.11$ & $-3.42\text{ to }-0.80$ \\
Without $V_t$ & $70.29\pm1.14$ & $-2.07$ & $-3.40\text{ to }-0.84$ \\
Without $1-H_t$ & $71.22\pm0.32$ & $-1.13$ & $-2.42\text{ to }+0.47$ \\
\bottomrule
\end{tabular}
\end{table*}

\begin{table}[t]
\caption{Selection audit pooled over 15 runs (135,000 selections per policy). $n_{\mathrm{eff}}$ is the entropy-effective number of transformation combinations. Risky denotes a combination containing elastic deformation, morphology, or erasing. Shares are percentages.}
\label{tab:selection_audit}
\centering
\scriptsize
\setlength{\tabcolsep}{6.5pt}
\renewcommand{\arraystretch}{1.5}
\begin{tabular}{@{}lrrrl@{}}
\toprule
Policy & Selections & $n_{\mathrm{eff}}$ & Risky & Top pair (share) \\
\midrule
ART & 135,000 & 5.71 & 2.05 & small rotation--small scaling \\
Uniform-28 & 135,000 & 28.00 & 64.19 & mild stroke noise--mild morphology \\
\bottomrule
\end{tabular}
\end{table}

\begin{table*}[bt]
\caption{ART-only CPU scaling. Latencies are warmed per-query median/p95 in ms. Memory is reported in MiB for one/five projection heads; temporary memory is the estimated sum of retained scoring arrays.}
\label{tab:art_systems}
\centering
\small
\setlength{\tabcolsep}{15pt}
\renewcommand{\arraystretch}{1.5}
\begin{tabular}{@{}rrrrrr@{}}
\toprule
IDs & Scars & Retrieval & Model incl. & Index 1/5 & Temp. est. \\
 & & med./p95 & med./p95 & MiB & MiB \\
\midrule
100 & 1,067 & 1.85/2.55 & 121.22/209.26 & 2.09/10.43 & 0.50 \\
250 & 2,171 & 4.64/5.58 & 123.65/225.46 & 4.26/21.22 & 1.03 \\
500 & 3,950 & 8.46/11.11 & 127.76/219.46 & 7.74/38.60 & 1.87 \\
1,000 & 7,663 & 11.82/14.78 & 130.66/230.79 & 15.03/74.89 & 3.62 \\
1,635 & 12,221 & 18.70/22.21 & 138.85/224.17 & 23.96/119.44 & 5.78 \\
\bottomrule
\end{tabular}
\end{table*}

\subsection{Confirmatory Transformed-Scar Retrieval}
\label{sec:confirmatory_results}

Table~\ref{tab:confirmatory_core} reports the locked five-method comparison
under the protocol in Section~\ref{sec:baselines_statistics}.

Table~\ref{tab:art_uniform_primary} reports ART at 71.40\% balanced Top-1 and uniform-28 at 68.83\%. Their paired ART-minus-uniform difference is 2.57 percentage points (95\% CI: 1.76 to 3.32). The prespecified strong-confirmation rule is met because the gain exceeds one percentage point and the paired interval excludes zero. The marginal method intervals overlap, but the prespecified inference uses the 15 within-fold--seed paired differences rather than overlap of marginal intervals.

The paired ART-minus-reference balanced differences were Fixed conservative augmentation: 1.01 (95\% CI: 0.01 to 2.10); RandAugment (N=2, M=9): 13.27 (95\% CI: 11.77 to 14.88); Conventional full augmentation: 11.09 (95\% CI: 9.50 to 12.51) percentage points.

Fixed conservative augmentation is retained as a strong domain-informed baseline: ART is higher in the balanced contrast by 1.01 (95\% CI: 0.01 to 2.10) percentage points, while ART has the larger mean in the elastic test set; the fixed policy has the larger mean in the safe-affine and mixed-transformation test sets. Because it jointly samples a manually restricted safe-affine policy rather than ART's 28 transformation combinations, it contextualizes performance but does not control for adaptive selection. Uniform-28 is the matched transformation-space control.

\subsection{Fixed Sketch-Level Retrieval}
\label{sec:sketch_level_results}

Table~\ref{tab:sketch_level_retrieval} reports the complete
sketch-to-sketch comparison. Among the baselines, CLIP has the highest
Top-1 accuracy at 18.18\%, while DINOv2 has the highest MRR at 0.2481. With
appearance alone, ART best-pair matching reaches 14.29\% Top-1, whereas
query-directed-mean aggregation reaches 24.68\%. Adding registered location
raises these values to 35.06\% and 55.84\%, respectively. Thus, with the same
location-aware pair score, requiring evidence from every query scar improves
Top-1 by 20.77 percentage points over retaining only the globally strongest
scar pair.

\subsection{Reward and Selection Diagnostics}
\label{sec:reward_diagnostics}

Table~\ref{tab:reward_ablation} reports the four leave-one-component-out variants. The removed weight is set to zero and the remaining weights are renormalized to sum to one. The full-reward seed-42 ART cells are reused as paired controls. The paired interval lies below zero when removing preservation $P_t$, consistency $C_t$, and useful variation $V_t$; those removals produce diagnostically resolved degradation under this protocol. Removing the harm-avoidance term $1-H_t$ produces an interval that crosses zero.

Table~\ref{tab:selection_audit} pools 135,000 selections for each policy. Let
$n_T$ be the pooled number of selections of combination $T\in\mathcal A$ and define
$p_T:=n_T/\sum_{U\in\mathcal A}n_U$. The entropy-effective number of combinations is
$n_{\mathrm{eff}}:=\exp[-\sum_{T\in\mathcal A}p_T\log p_T]$, with
$0\log0:=0$. ART uses 5.71 effective pairs versus 28.00 for uniform-28. ART
assigns 2.05\% of selections to pairs containing elastic deformation,
morphology, or erasing, compared with 64.19\% for uniform-28. The latter is
close to the transformation-space expectation $18/28=64.29\%$ because 18 of the 28 combinations
contain at least one such transformation. ART's three most selected affine combinations
account for 80.62\% of all selections; small rotation--small scaling alone
accounts for 41.00\%. This concentration reflects the complete
controller---reward updates, initialization, UCB exploration, and the fixed
safety adjustment---and therefore is not attributable to reward alone. The uniform policy
uses an independent random stream over the identical audited transformation space, and
logged rewards never alter its trace.

\subsection{ART Retrieval-System Scaling}
\label{sec:systems_results}

Table~\ref{tab:art_systems} characterizes the ART retrieval path as a system diagnostic rather than an accuracy or identity-generalization experiment. The full database contains 12,221 scars and takes 138.79 s to index. Measurements use Apple M1 Pro, PyTorch 2.8.0, and CPU execution throughout.

From the smallest to the full database, scar count increases by 11.45$\times$ (1,067 to 12,221), while median retrieval-only latency increases by 10.09$\times$ (1.85 to 18.70 ms),
indicating roughly proportional exact-search scaling over the measured range. Index memory and temporary scoring memory increase linearly with scar count.
Model-inclusive median latency rises only from 121.22 to 138.85 ms because query encoding dominates the loaded retrieval path.
These descriptive measurements establish computational feasibility.

The timing protocol excludes one-time model loading, warms exact retrieval with five queries per database size, and warms the embedding model once.

\section{Conclusion}
\label{sec:conclusion}

This paper introduces Augmentation with Reinforcement Transformation (ART) for low textured scientific sketch image retrieval using Florida manatee sketches. ART uses a label-free reward for structural preservation, feature consistency, useful variation, and transformation harm to guide reward-based selection over pairwise combinations of eight image transformations. ART first caches its reward-guided views and their frozen-backbone features; VICReg then jointly optimizes the retained representation head and a training-only expander. The VICReg loss does not feed back to ART, and the expander is discarded before the retained encoder maps query and database scars into a shared embedding space for sketch ranking.

Across five locked folds and three optimization seeds, ART outperforms uniform sampling over the identical 28-combination transformation space on balanced Top-1. The primary paired comparison satisfies the prespecified strong-confirmation rule, supporting the complete ART controller over uniform transformation sampling on this fixed transformed-scar protocol. Fixed conservative augmentation remains a strong domain-informed baseline rather than the matched transformation-space control.

On the separate fixed sketch-level evaluation, location-aware directed-mean
ART matching achieves 55.84\% Top-1 accuracy, compared with 18.18\% for the
strongest frozen whole-image baseline and 35.06\% for location-aware best-pair
matching. This shows the value of combining registered location with evidence
from every query scar on this split.

Reward ablations show degradation after removing preservation, consistency, and useful variation, while removal of harm avoidance is unresolved. The transformation-frequency audit nevertheless makes the learned selection behavior observable and distinguishes it from the near-uniform control.

The system evaluation scales exact retrieval to 1,635 database identities and 12,221 scars, with a retrieval-only median/p95 of 18.70/22.21 ms per query. These measurements establish computational feasibility, not superiority on unseen identities. The evidential scope remains transformed instances and fixed cross-sketch matches from one manatee sketch database. Future work should enlarge a prospectively locked identity-disjoint cohort, test temporal and cross-database transfer, and compare against a broader set of automatic augmentation methods.

\section*{Acknowledgment}
This work has been supported in part by the U.S. National Science Foundation (NSF) under the Partnership to Advance Conservation Science and Practice (PACSP) program Grant Nos.\, IOS-2430224, IOS-2446522, and by the Paul G. Allen Family Foundation.

\bibliographystyle{IEEEtran}
\bibliography{references}

\end{document}